\documentclass{article}
\usepackage{spconf,amsmath,graphicx,amsfonts,url}

\title{Zero-shot Imitation Policy via Search in Demonstration Dataset}
\name{Federico Malato\sthanks{Equal contribution}\textsuperscript{1}, Florian Leopold\textsuperscript{*2}, Andrew Melnik\textsuperscript{2},  Ville Hautam\"aki\textsuperscript{1}}
\address{\textsuperscript{1}School of Computing, University of Eastern Finland, Finland\\
\textsuperscript{2}Bielefeld University, Germany\\
federico.malato@uef.fi}

\begin{document}
\ninept
\maketitle
\begin{abstract}
Behavioral cloning uses a dataset of demonstrations to learn a policy. To overcome computationally expensive training procedures and address the policy adaptation problem, we propose to use latent spaces of pre-trained foundation models to index a demonstration dataset, instantly access similar relevant experiences, and copy behavior from these situations. Actions from a selected similar situation can be performed by the agent until representations of the agent's current situation and the selected experience diverge in the latent space. Thus, we formulate our control problem as a dynamic search problem over a dataset of experts' demonstrations. We test our approach on BASALT MineRL-dataset in the latent representation of a Video Pre-Training model. We compare our model to state-of-the-art, Imitation Learning-based Minecraft agents. Our approach can effectively recover meaningful demonstrations and show human-like behavior of an agent in the Minecraft environment in a wide variety of scenarios. Experimental results reveal that performance of our search-based approach clearly wins in terms of accuracy and perceptual evaluation over learning-based models.
\end{abstract}
\section{Introduction}
\label{sec:intro}
\textit{Imitation Learning} (IL)~\cite{Schaal1996} and \textit{Reinforcement Learning} (RL)~\cite{SuttonBarto1998,schilling2019approach,bach2020learn,schilling2021decentralized} are commonly used methods to train agents to perform tasks in simulated and real environments. Despite the extensive research in these fields, a number of persistent challenges still hold. Among them, significant computational costs and lack of zero-shot adaptability are the most prominent. Recent works leverage large language~\cite{ouyang2022training, lifshitz2023steve1, wang2023voyager} and vision models~\cite{oquab2023dinov2, rana2023contrastive} to achieve few-shot adaptability. Still, such models are computationally expensive to train. Therefore, exploring alternative approaches to IL and RL methods for control problems can potentially address these challenges and yield advantages in specific application domains.

{\em Behavioral Cloning} (BC)~\cite{TorabiStone2018} has been successfully applied to practical control problems ranging from autonomous driving~\cite{Saksena2019, Samak2021, beohar2022planning} to playing video games~\cite{KanervistoHautamaki2020, KanervistoPussinen2020, Vinyals2019}. Despite being tremendously popular due to its simplicity, BC suffers from a range of problems such as distributional shift and causal confusion~\cite{DeHaan2019,Russell2019}. Such limitations have been addressed with \textit{inverse reinforcement learning}~\cite{ng2000} (IRL) or \textit{generative adversarial imitation learning}~\cite{ho2016generative} (GAIL), which on the other hand tend to be computationally expensive and hard to train~\cite{arora2020survey,adams2022survey}. Moreover, in case of complex scenarios, such methods might struggle in learning a suitable strategy~\cite{Russell2019, ng2000}.

Previous literature has addressed agents adaptability problems by combining time-step-wise kNN search in latent space with Locally Weighted Regression~\cite{Cleveland1988} for action selection~\cite{pari2021surprising,haldar2023teach,bahl2023affordances}. Nonetheless, studies in this direction have been conducted only in controlled robotic environments with continuous actions. Additionally, in these environments a reward signal is either used, such as in~\cite{haldar2023teach,bahl2023affordances}, or can be inferred, for instance in~\cite{pari2021surprising}. Our work extends the applicability of search-based methods to discrete-action domains, and validates them in open-ended environment. Moreover, we demonstrate zero-shot adaptation for such methods.

To this end, we introduce \textit{Zero-shot Imitation Policy} (ZIP), a search-based approach to imitation learning that instantly adapts to new tasks in complex, discrete domains. ZIP encodes all experts' trajectories into a reasonable latent space. We define the concept of {\em situation}, a self-contained trajectory that includes both states and corresponding actions. We could think about situation as a {\em shot} in a movie. ZIP compares the current situation with the previous experiences encoded in latent space, searches for the closest one and executes the actions in it until the situation changes. This allows the policy to adapt to changes in the environment by just collecting new expert trajectories in that configuration. Perceptual experiments show that the proposed method clearly outperforms other agents except purely scripted agents. 

\section{Methods}
\label{sec:methods}
The present study is motivated by the MineRL BASALT 2022 challenge~\cite{ShahDragan2021, Milani2023}. In the challenge, an agent must solve four tasks: find a cave, build an animal pen, build a village house, and make a waterfall~\cite{ShahDragan2021, Milani2023}. For these tasks, no reward function is provided. To solve the tasks, participants are provided with a dataset of demonstrations, each showing a human expert solving one of the tasks. 

The core idea behind our approach is to reformulate the control problem as a search problem on the experts' demonstrations. 

\begin{figure}[h]
    \centering
    \includegraphics[width=0.95\columnwidth]{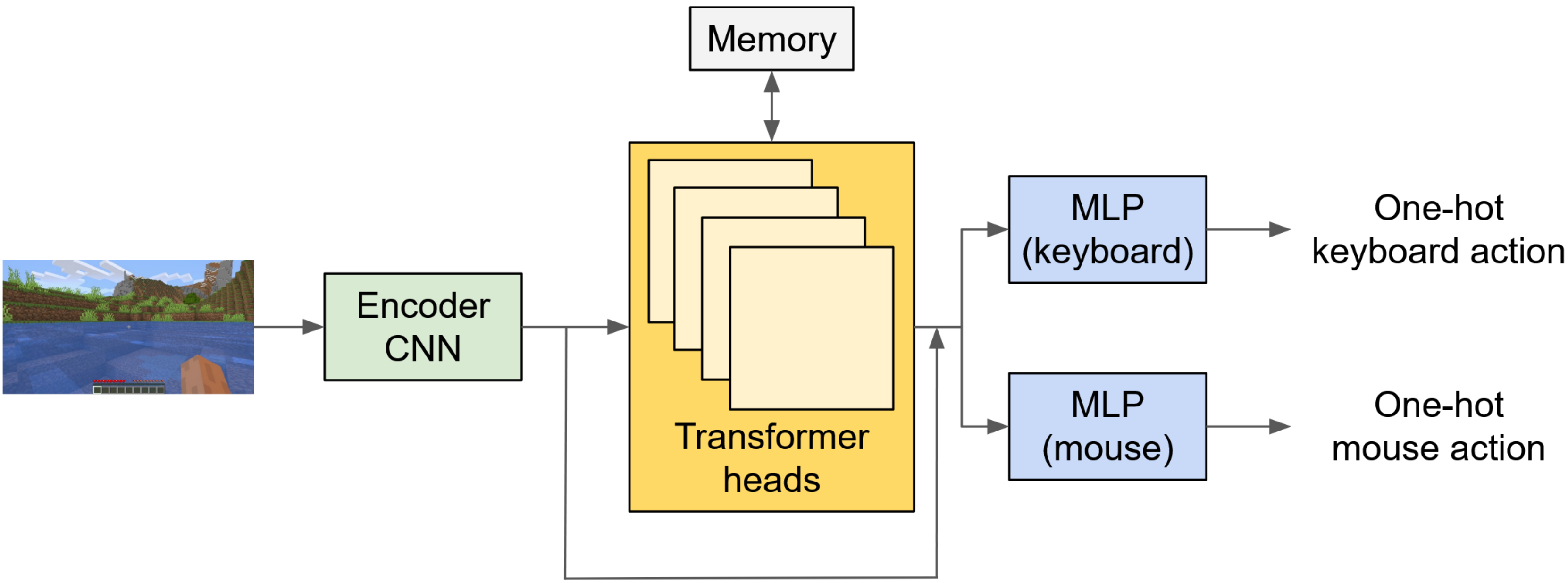}
    \caption{A scheme of the VPT model used in this study. An image input is encoded with an IMPALA CNN and passed through four transformer heads. Then, two MLP heads predict a keyboard and a mouse action respectively.}
    \label{fig:vpt_architecture}
    \vspace{-0.1in}
\end{figure}

To keep the search computationally feasible, we take advantage of large \textit{pre-trained} models of the problem domain.
Recently proposed {\em Video Pre-Training} (VPT) model~\cite{baker2022video} (see Figure \ref{fig:vpt_architecture}) is a foundation model for BC trained on 70k+ hours of video content scraped from the internet~\cite{baker2022video}. VPT is built on an IMPALA~\cite{EspeholtKavukcuoglu2018} convolutional neural network (CNN) backbone. The CNN maps an image input to a 1024-dimensional feature vector. VPT generates a batch of 129 vectors and forward them to four transformer blocks. Each block is linked to a memory block containing the last 128 frames. At the end of the transformer pipeline, only the last frame is retained and forwarded to two heads based on Multilayer Perceptrons (MLPs). One head predicts a keyboard action, while the second head predicts a computer mouse action.

We use a pre-trained VPT model~\cite{baker2022video} to encode situations in latent space. The pre-trained version of the model used in this study is available at the official GitHub repository~\cite{VPTGithub}. The repository features three foundation models, namely \textit{1x}, \textit{2x} and \textit{3x}. The backbone of the three models is the same, and they differ only for the weights width~\cite{VPTGithub}.

\begin{figure}[t!]
    \centering
    \includegraphics[width=0.95\columnwidth]{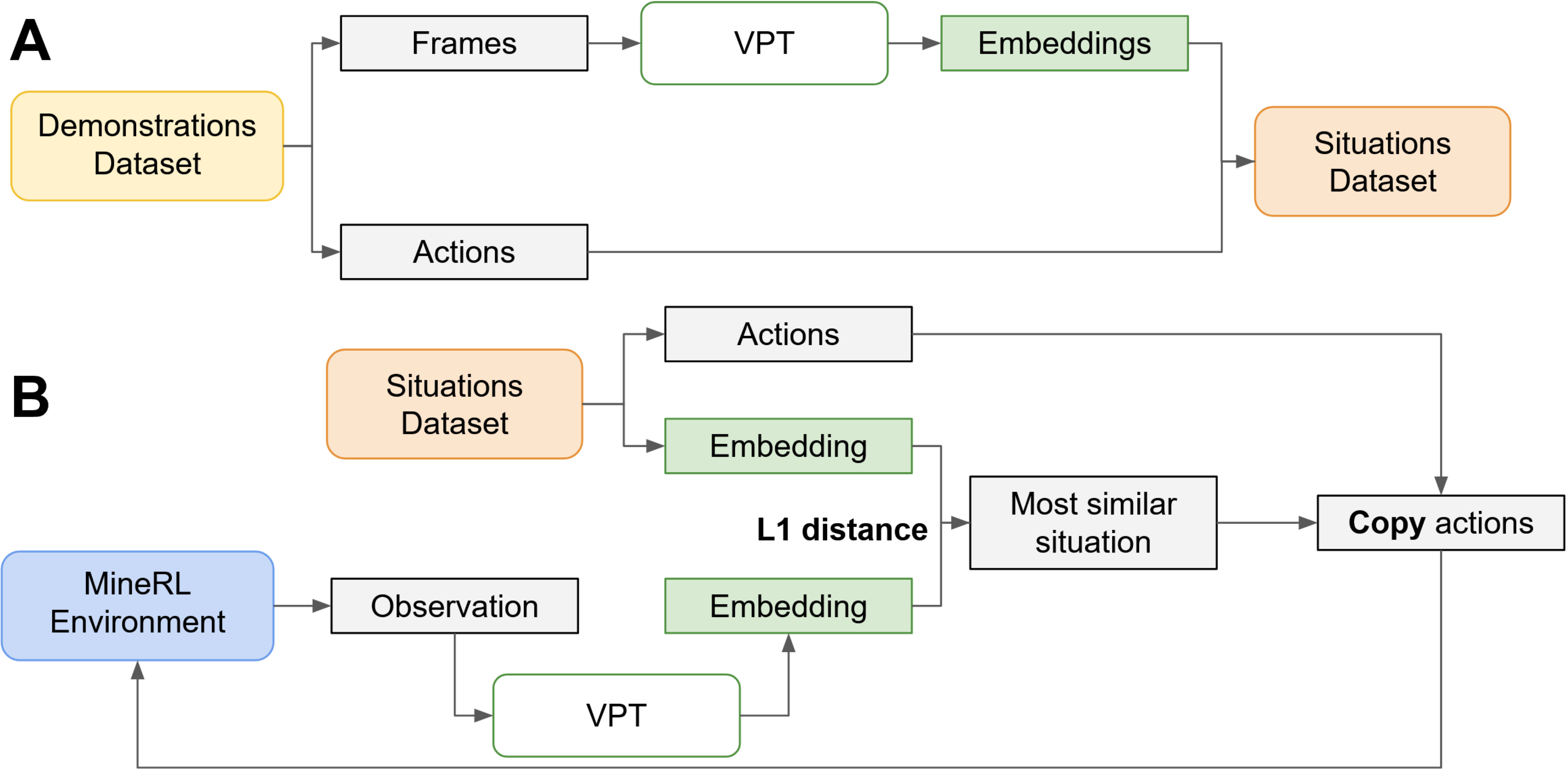}
    \caption{Our approach. (\textbf{A}) \textit{Latent space generation:} trajectories are extracted from the demonstration dataset. Frames are encoded through a provided VPT model, and paired with the corresponding actions. (\textbf{B}) \textit{Evaluation loop:} at each time-step, the new observation is forwarded to the same VPT model. Then, L1 distance across current and reference embeddings is computed and the most similar situation is found. ZIP acts in the environment following the actions of the selected reference situation.}
    \label{fig:approach}
\end{figure}

\begin{figure}[h]
    \centering
    \includegraphics[width=0.9\columnwidth]{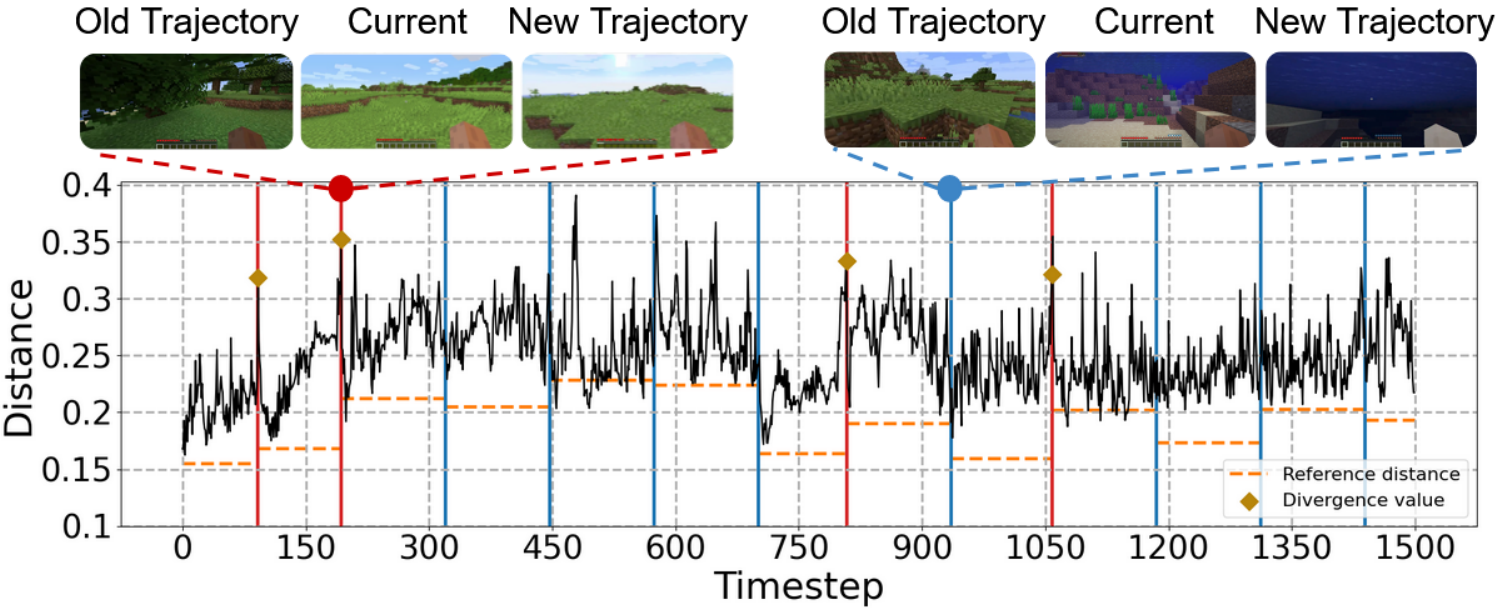}
    \caption{An example of the search mechanism. At each time-step, we keep track of the distance between current and reference embedding. Whenever the distance overcomes a threshold, a divergence-based search (red line) selects a new reference embedding; if the agent follows a threshold for too long, a time-based search (blue line) is triggered. For each segment of the episode a yellow, dashed line indicates the value of the reference distance. A brown diamond corresponding to each red line shows the distance value that triggered the new search.}
    \label{fig:divergence}
    \vspace{-0.1in}
\end{figure}

\subsection{Zero-shot Imitation Policy}

ZIP solves a control problem by retrieving relevant past experiences from experts' demonstrations. In this context, we define a \textit{situation} as a set of consecutive observation-action pairs $\{(o_{t}, a_{t}), \dots, (o_{t+\tau}, a_{t+\tau})\}$, $\tau \in \mathbb{N}$. 

We illustrate our approach in Figure \ref{fig:approach}. We use VPT to extract all the embeddings from an arbitrarily chosen subset $\mathcal{S}$ of the available demonstrations dataset $\mathcal{D}$, $\mathcal{S} \subseteq \mathcal{D}$. $\mathcal{S}$ constitutes the $d$-dimensional latent space that ZIP searches.  Moreover, we assume that expert has completed the given task (find cave, build waterfall, etc), so optimality assumption in IL is satisfied~\cite{Russell2019}. 

During testing (see Figure \ref{fig:approach}B), we pass the current observation through VPT. Then, ZIP selects the most similar reference trajectory embedding in the latent space, according to their L1 distance with the current observation. Finally, it copies the actions of the selected reference situation. At each time-step, we shift the current and reference situations in time and recompute their distance. When situations diverge over time, the approach performs a new \textit{divergence-triggered} search (red lines in Figure \ref{fig:divergence}). Additionally, if the reference embedding is followed for more than $n$ time-steps, ZIP performs a new \textit{time-triggered} search (blue lines in Figure \ref{fig:divergence}).

\begin{figure}[t!]
    \centering
    \includegraphics[width=0.66\columnwidth]{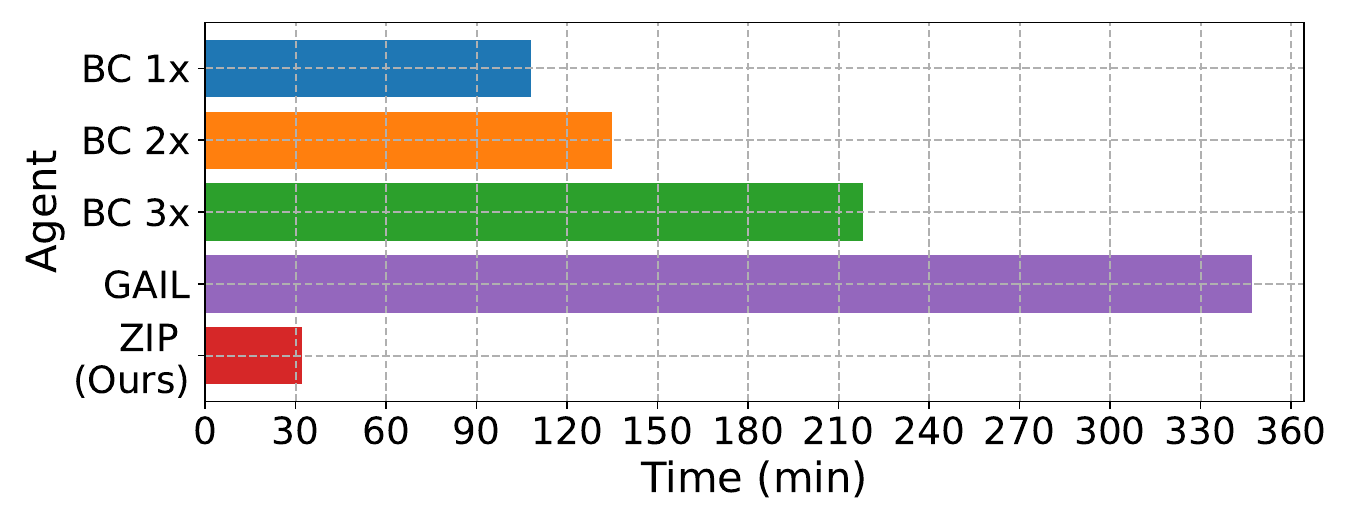}
    \caption{Time needed to train each agent on 100 trajectories on the FindCave task. In the case of BC models, the training procedure consists of fine-tuning a pre-trained VPT model. For ZIP, training means encoding a subset of trajectories through the reference version of VPT. All models have been trained on a single Tesla T4 GPU.}
    \label{fig:training_time}
\end{figure}

\section{Experiments}
\label{sec:experiments}
The full FindCave dataset from the MineRL BASALT competition~\cite{ShahDragan2021} consists of 5466 experts' trajectories demonstrating the task, or around 150GB of data. For each frame of an episode, only the unprocessed RGB image and the corresponding action are available. That is, an episode is a set of image-action pairs. No reward signal is available. Similarly, no measure of performance is provided. In our study we considered only the first $100$ trajectories, a small fraction of the available data.

For our comparison, we fine-tuned the three VPT-based, pre-trained foundation models introduced in Section \ref{sec:methods}. Additionally, we compare ZIP to GAIL~\cite{ho2016generative}, a state-of-the-art IL algorithm. We train GAIL from scratch until convergence for almost $6$ hours on the same subset of data, using VPT as encoder. While ZIP does not require any training, we refer to training as the process of encoding the experts' trajectories through VPT (see Figure \ref{fig:approach}A). A comparison of training times for our tested agents is shown in Figure \ref{fig:training_time}. Notably, setting up ZIP (32 minutes) requires much less time than fine-tuning any VPT-based BC agent (respectively, 108, 135 and 212 minutes), or training GAIL (347 minutes).

We perform two sets of experiments. First, we analyse the results of the perceptual evaluation from BASALT \cite{Milani2023}, where human contractors grade pairs of videos. Second, we approximate a quantitative evaluation for the FindCave task from the BASALT suite. We decided to evaluate only FindCave because, despite the other tasks, we found a sufficiently reliable way of establishing a ground truth for its completion. To this end, we train a simple binary classifier to detect the presence of a cave. Our classifier is composed by four convolutional layers and two fully-connected layers. The network has been trained on a dataset of cave/non-cave frames, manually extracted from the original FindCave data, and achieved a validation accuracy of $97.89\%$ over it.

We test our agents on three-minutes-long episodes, similarly to the official BASALT rules. We apply minimal changes to the terminal condition of an episode, to ease the evaluation procedure while keeping the performance evaluation intact. First, we disable the terminal action \textit{'ESC'} in all agents. We support this decision by highlighting that terminal actions constitute a minimal fraction of the training examples, and a BC agent would likely ignore it~\cite{TorabiStone2018}. Second, the BASALT competition considers an episode to be successful whenever an agent performs the terminal action while being inside a cave. Instead, we consider an episode successful whenever an agent spends more than five seconds (that is, $100$ consecutive cave frames) in a cave. We justify this choice by considering that a time threshold keeps the original evaluation criterion intact, while accounting for false positives such as random cave sightings.

As in the evaluation process of the BASALT competition, we test our agent on twenty seed values. Since the official values used in BASALT evaluation are not publicly available, we have selected each seed manually to ensure the presence of caves. We repeat three runs over our set of seeds.

Our approach relies on two parameters, \textit{maximum steps} and \textit{divergence scaling factor}, regulating the frequencies of the two types of search. \textit{Maximum steps} regulates the maximum number of consecutive actions that an agent can use from the same trajectory, before triggering a new time-based search. On the other hand, \textit{divergence scaling factor} determines when a new divergence-based search is triggered, based on how much the distance between current and reference embeddings has increased compared to the last search. For the perceptual experiment, we have set $\textit{maximum steps}=128$ and $\textit{divergence scaling factor}=2.0$. We perform an ablation study over these hyperparameters, selecting nine values for each, centered around our reference value. We test our agent for three runs of ten episodes each, using a fixed seed.

Finally, we visualise and analyse the latent space generated from VPT using a t-SNE plot. For visual clarity, we encode only $10$ of the $100$ trajectories used in our experiments. We differentiate trajectories by marking their points with different colors. Additionally, we distinguish exploration frames from cave frames, and analyse their distribution through the space.

\section{Results}
\label{sec:results}
We report both the perceptual and quantitative evaluations for our agent. Notably, perceptual results have been obtained by asking anonymous human contractors to compare randomly selected pairs of agents. For each comparison, agents have been evaluated on \textit{human-likeliness} and \textit{success}. 
\subsection{Perceptual evaluation}
\begin{table}[!t]
\centering
\small
\caption{Top-5 ranking of the NeurIPS BASALT 2022 competition~\protect\cite{Milani2023}. Below, the TrueSkill~\protect\cite{minka2018trueskill} scores for two human expert players, a BC baseline, and a random agent.}
{\begin{tabular}{@{}cccccc@{}}
\hline \\[-6pt]
\textbf{Team} & Find & Make & Build & Build & \textbf{Average} \\[1pt]
 & Cave & Waterfall & Pen & House &  \\[1pt]
 \hline \\[-6pt]
GoUp & $0.31$ & $\textbf{1.21}$ & \textbf{0.28} & \textbf{1.11} & \textbf{0.73} \\[1pt]
\textbf{ZIP} & \textbf{0.56} & -0.10 & 0.02 & 0.04 & \textbf{0.13} \\ [1pt]
voggite & 0.21 & 0.43 & -0.20 & -0.18 & \textbf{0.06} \\[1pt]
JustATry & -0.31 & -0.02 & -0.15 & -0.14 & \textbf{-0.15} \\[1pt]
TheRealMiners & 0.07 & -0.03 & -0.28 & -0.38 & \textbf{-0.16} \\[1pt]
\hline \\[-6pt]
Human2 & 2.52 & 2.42 & 2.46 & 2.34 & \textbf{2.43} \\[1pt]
Human1 & 1.94 & 1.94 & 2.52 & 2.28 & \textbf{2.17} \\[1pt] \hline \\[-6pt]
BC-Baseline & -0.43 & -0.23 & -0.19 & -0.42 & \textbf{-0.32} \\[1pt]
Random & -1.80 & -1.29 & -1.14 & -1.16 & \textbf{-1.35} \\[1pt] \hline \\[-9pt]
\end{tabular}
}
\label{tab:results}
\vspace{-0.2in}
\end{table}
The organization committee of the competition ranked the agents using the TrueSkill~\cite{minka2018trueskill} ranking system, which is widely used in the Microsoft online gaming ecosystem. Given a set of competitors, the system uses Bayesian inference to compute an ELO-like score, according to the match history of each competitor.

Our proposed ZIP agent was overall ranked second place in the challenge. The results of our agent are described in Table~\ref{tab:results}. The first place was awarded to team \textit{GoUp}, who leveraged detection methods and human knowledge of the task and combined them with scripting~\cite{Milani2023}. Notably, all the other learning-based methods achieved lower performance than ZIP in three out of four tasks. Additionally, our method was awarded with $2$ out of $5$ research innovation prizes from the organizing committee.

\begin{figure}[!t]
    \centering
    \includegraphics[width=0.9\columnwidth]{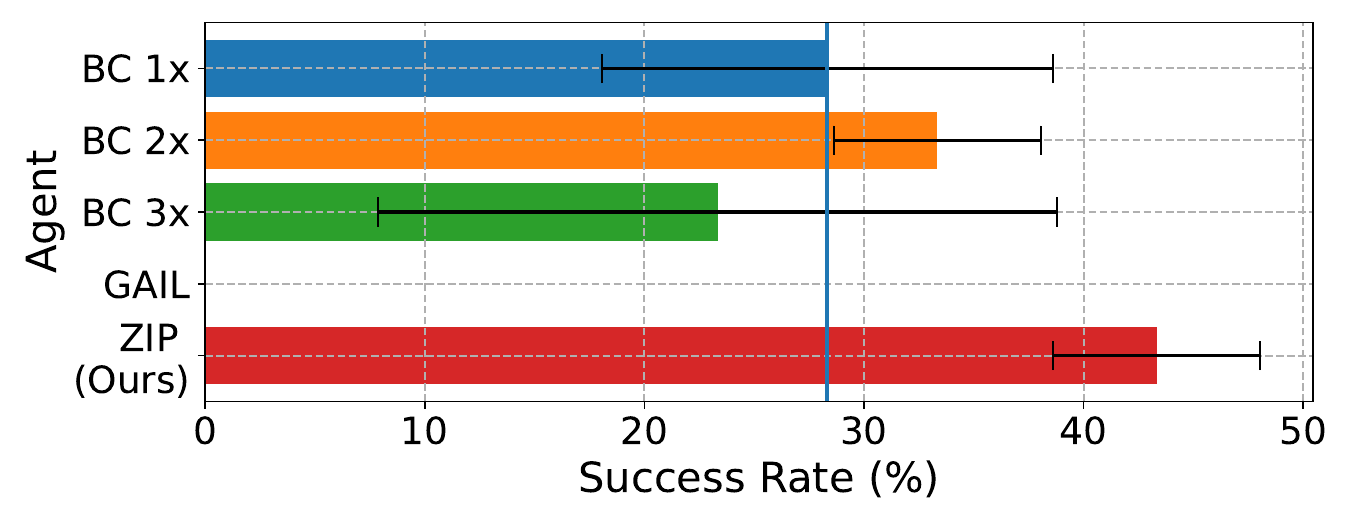}
    \caption{Average success rate for tested models on the FindCave task. Each agent has been evaluated on a batch of $20$ seeds. Each run has been repeated three times. Baseline model is highlighted with a vertical blue line.}
    \label{fig:cave_results}
    \vspace{-0.3in}
\end{figure}

\subsection{Quantitative results}
We report success rate for the tested models on the FindCave task in Figure \ref{fig:cave_results}. ZIP has obtained the best performance, being able to complete the task $43.32 \pm 4.71\%$ of the times. Notably, its worst performance on our repeated trials has reported a $38.59\%$ success rate, comparable to the best run among all the BC agents ($38.88\%$, \textit{BC 3x}). 

GAIL was never able to complete the task. We have compared our setup with~\cite{ho2016generative} to account for errors in the hyperparameters values. We have trained GAIL for $6$, $12$, $18$ and $22$ hours. We have observed a saturation of policy and discriminator losses after $6$ hours. By watching some videos of the trained policy playing the game, we noticed an improvement over a random agent. Still, the trained agent had difficulties in completing basic actions consistently.

On the contrary, VPT-based BC models have been pre-trained on huge amount of data scraped from the internet. As a consequence, fine-tuning them led to quite successful performance. More specifically, \textit{BC 1x} succeeded (on average) $28.33 \pm 10.27\%$ of the time, while \textit{BC 2x} reached $33.4 \pm 4.71\%$ of success rate. Perhaps surprisingly, \textit{BC 3x} completed the task only $23.5 \pm 15.46\%$ of times. We explain this result by observing that our tested BC agents have generally higher variance than ZIP. Thus, we believe that \textit{BC 3x} only suffered from poor choices of actions.

\begin{figure}[!t]
    \centering
    \includegraphics[width=0.9\columnwidth]{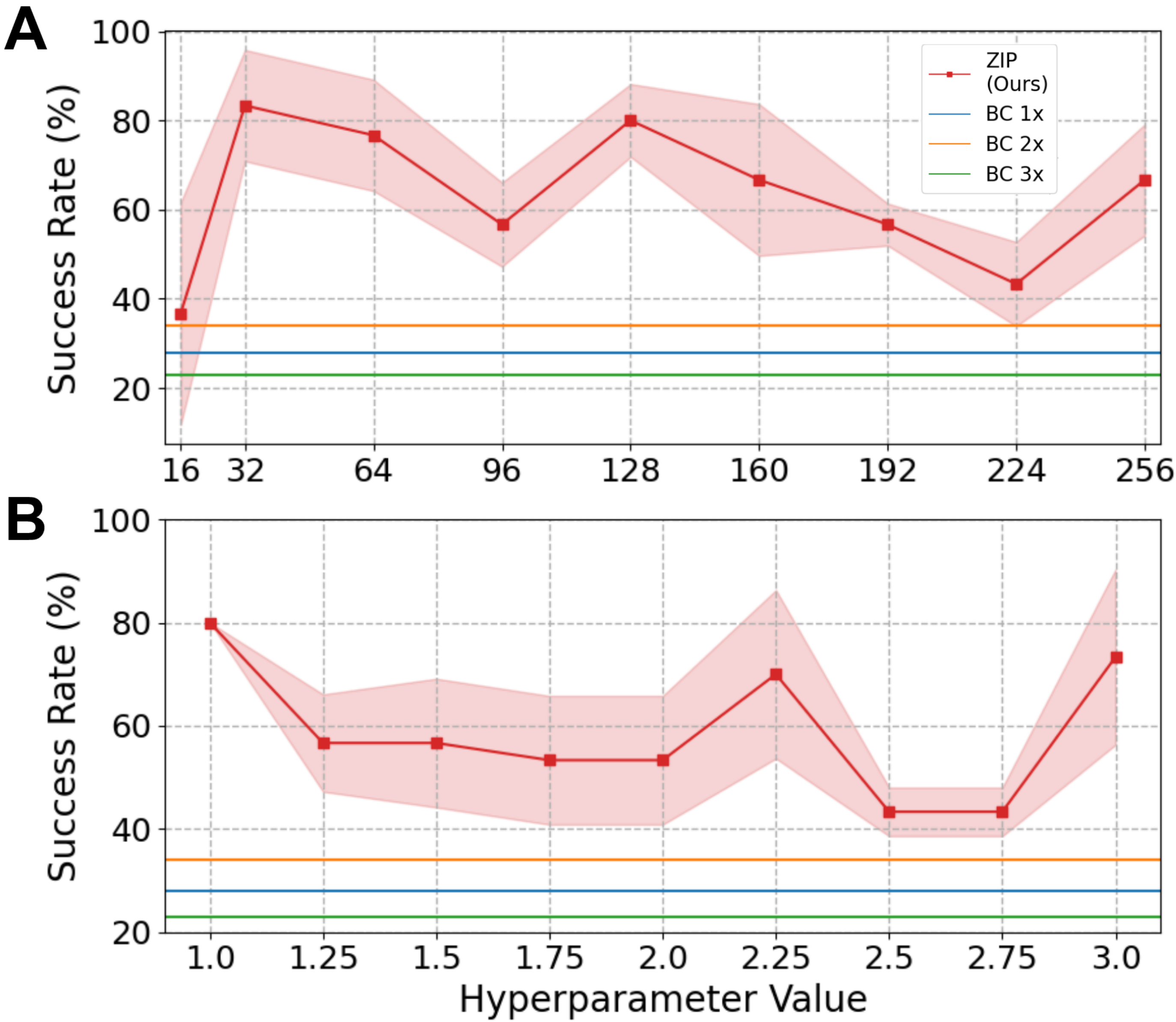}
    \caption{Ablation study over the hyperparameters of our proposed method. \textbf{(A)} Maximum number of time-steps following the same trajectory. \textit{X-axis}: values of the time-steps threshold; \textbf{(B)} Divergence scaling factor. \textit{X-axis}: divergence factor values. For visual clarity, only average performance of the BC models is reported.}
    \label{fig:ablation}
    \vspace{-0.2in}
\end{figure}

\subsection{Ablation study}
In Figure \ref{fig:ablation} we report the results of the ablation study conducted over the hyperparameters of ZIP. While we were not able to identify a precise pattern, it is clear that some values yield better results than others. For instance, \textit{maximum steps} (Figure \ref{fig:ablation}A) seems to improve ZIP's performance when its value is between $32$ and $128$. This results confirms what we have found empirically, that is, $128$ is a good candidate.

As for \textit{divergence scaling factor}, it appears that either $1.0$, $2.25$ and $3.0$ are good choices. In particular, $1.0$ yields no variance, suggesting very consistent performance. In our reference implementation, we have used a value of $2.0$, which does not seem to be competitive with other values.

Following the results of the ablation study, we ran again the quantitative experiment, changing only the hyperparameters values to $\textit{maximum steps}=32$ and $\textit{divergence scaling factor}=1.0$. We found that ZIP replicates the results obtained by our reference setup almost perfectly ($43.3 \pm 6.24\%$). We justify this result by considering that the ablation study has been conducted on a fixed seed, while the quantitative study uses variable seed values. Therefore, in a complex environment such as Minecraft, changing conditions can lead to substantially different results. Nonetheless, it is notable how our agent was able to keep the same performance despite the changes.

\begin{figure}[!t]
    \centering
    \includegraphics[width=\columnwidth]{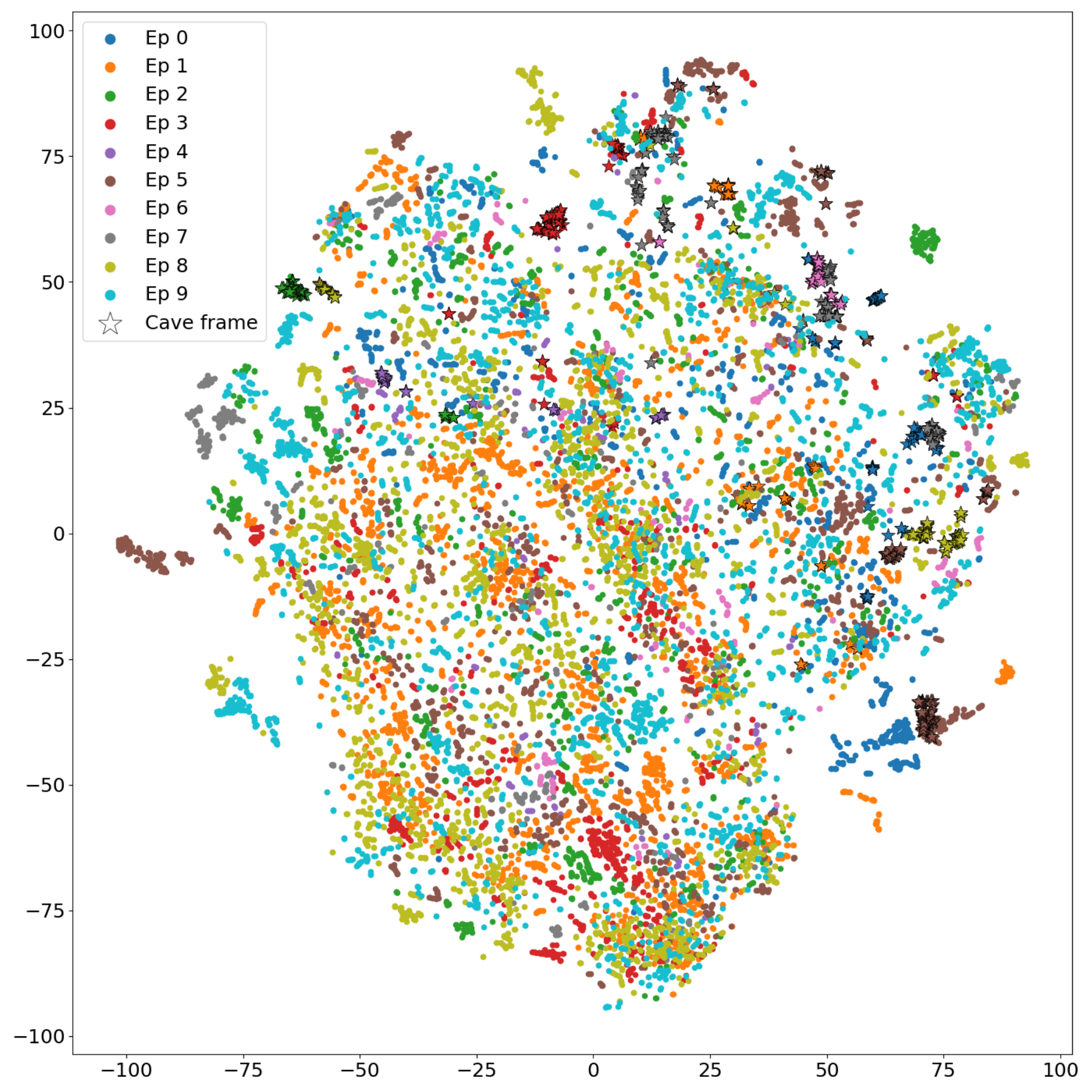}
    \caption{Example of latent space generated by VPT. The example shows only $10$ out of $100$ trajectories for visual clarity. Points belonging to the same trajectory are marked with the same color. We highlight cave frames using a star-shaped marker.}
    \label{fig:latent_space}
    \vspace{-0.2in}
\end{figure}
\subsection{Latent space visualisation}
In Figure \ref{fig:latent_space} we show an example of latent space generated through the VPT encoding. Each point represent one frame belonging to a specific trajectory, separated by color. Star-shaped points refer to cave frames of a specific trajectory.

Clusters of points are clearly distinguished. In general, clusters are heterogeneous, even though some smaller clusters at the margins of the space are formed by points of one or two demonstrations. Those smaller clusters represent peculiar and rare situations that happened only once or twice, for instance, spawning in a desert area.

Figure \ref{fig:latent_space} also shows that cave frames are concentrated on the outskirts of the space, mostly on the right side. Interestingly, cave frames belonging to one trajectory tend to be close to each other, while being well separated from cave frames of another trajectory. We suggest that the distribution of those frames might allow for more refined strategies related to goal conditioning, or to navigate the latent space.

\section{Conclusions}
We present Zero-shot Imitation Policy (ZIP), a search-based approach to behavioral cloning that efficiently adapts to new tasks, leveraging pretrained models in complex domains. Our experiments show that ZIP is a robust alternative to imitation learning methods, while requiring very short amount of time to be set up.

ZIP is mostly limited by the size of the latent space and the quality of data used. Future works could explore the usage of external tools for better data compression~\cite{faiss} or measures for ranking available data according to their relevance.

\section{Acknowledgements}
We thank Research Council Finland (project number 350093)  and Detuscher Akademischer Austauschdienst (DAAD) for supporting the collaboration between our groups.
\bibliographystyle{IEEEbib}
\bibliography{bibliography}

\begin{thebibliography}{10}

\bibitem{Schaal1996}
Stefan Schaal,
\newblock ``Learning from demonstration,''
\newblock in {\em Advances in Neural Information Processing Systems}, M.C.
  Mozer, M.~Jordan, and T.~Petsche, Eds. 1996, vol.~9, MIT Press.

\bibitem{SuttonBarto1998}
Richard~S. Sutton and Andrew~G. Barto,
\newblock {\em Reinforcement Learning: An Introduction},
\newblock The MIT Press, second edition, 2018.

\bibitem{schilling2019approach}
Malte Schilling and al.,
\newblock ``An approach to hierarchical deep reinforcement learning for a
  decentralized walking control architecture,''
\newblock in {\em Biologically Inspired Cognitive Architectures 2018:
  Proceedings of the Ninth Annual Meeting of the BICA Society}. Springer, 2019,
  pp. 272--282.

\bibitem{bach2020learn}
Nicolas Bach and al.,
\newblock ``Learn to move through a combination of policy gradient algorithms:
  Ddpg, d4pg, and td3,''
\newblock in {\em International Conference on Machine Learning, Optimization,
  and Data Science}. Springer, 2020, pp. 631--644.

\bibitem{schilling2021decentralized}
Malte Schilling and al.,
\newblock ``Decentralized control and local information for robust and adaptive
  decentralized deep reinforcement learning,''
\newblock {\em Neural Networks}, vol. 144, pp. 699--725, 2021.

\bibitem{ouyang2022training}
Long Ouyang and al.,
\newblock ``Training language models to follow instructions with human
  feedback,''
\newblock {\em Advances in Neural Information Processing Systems}, vol. 35, pp.
  27730--27744, 2022.

\bibitem{lifshitz2023steve1}
Shalev Lifshitz, Keiran Paster, Harris Chan, Jimmy Ba, and Sheila McIlraith,
\newblock ``Steve-1: A generative model for text-to-behavior in minecraft,''
  2023.

\bibitem{wang2023voyager}
Guanzhi Wang, Yuqi Xie, Yunfan Jiang, Ajay Mandlekar, Chaowei Xiao, Yuke Zhu,
  Linxi Fan, and Anima Anandkumar,
\newblock ``Voyager: An open-ended embodied agent with large language models,''
  2023.

\bibitem{oquab2023dinov2}
Maxime Oquab and al.,
\newblock ``Dinov2: Learning robust visual features without supervision,''
\newblock {\em arXiv preprint arXiv:2304.07193}, 2023.

\bibitem{rana2023contrastive}
Krishan Rana and al.,
\newblock ``Contrastive language, action, and state pre-training for robot
  learning,''
\newblock {\em arXiv preprint arXiv:2304.10782}, 2023.

\bibitem{TorabiStone2018}
Faraz Torabi and al.,
\newblock ``{Behavioral cloning from observation},''
\newblock in {\em IJCAI International Joint Conference on Artificial
  Intelligence}, 2018, vol. 2018-July.

\bibitem{Saksena2019}
Saumya~Kumaar Saksena and al.,
\newblock ``Towards behavioural cloning for autonomous driving,''
\newblock 2019.

\bibitem{Samak2021}
Tanmay~Vilas Samak and al.,
\newblock ``Robust behavioral cloning for autonomous vehicles using end-to-end
  imitation learning,''
\newblock {\em SAE International Journal of Connected and Automated Vehicles},
  vol. 4, 2021.

\bibitem{beohar2022planning}
Shivansh Beohar and al.,
\newblock ``Planning with rl and episodic-memory behavioral priors,''
\newblock {\em arXiv preprint arXiv:2207.01845}, 2022.

\bibitem{KanervistoHautamaki2020}
Anssi Kanervisto and al.,
\newblock ``Playing minecraft with behavioural cloning,''
\newblock {\em CoRR}, vol. abs/2005.03374, 2020.

\bibitem{KanervistoPussinen2020}
Anssi Kanervisto and al.,
\newblock ``{Benchmarking End-to-End Behavioural Cloning on Video Games},''
\newblock in {\em IEEE Conference on Computatonal Intelligence and Games, CIG},
  2020, vol. 2020-August.

\bibitem{Vinyals2019}
Oriol~Vinyals Vinyals and al.,
\newblock ``Grandmaster level in starcraft ii using multi-agent reinforcement
  learning,''
\newblock {\em Nature}, vol. 575, 2019.

\bibitem{DeHaan2019}
Pim de~Haan, Dinesh Jayaraman, and Sergey Levine,
\newblock ``Causal confusion in imitation learning,''
\newblock 2019, vol.~32.

\bibitem{Russell2019}
Stuart Russell,
\newblock {\em Human Compatible},
\newblock Penguin, 2019.

\bibitem{ng2000}
Andrew~Y. Ng and Stuart~J. Russell,
\newblock ``Algorithms for inverse reinforcement learning,''
\newblock in {\em Proceedings of the Seventeenth International Conference on
  Machine Learning}, San Francisco, CA, USA, 2000, ICML '00, p. 663–670,
  Morgan Kaufmann Publishers Inc.

\bibitem{ho2016generative}
Jonathan Ho and al.,
\newblock ``Generative adversarial imitation learning,'' 2016.

\bibitem{arora2020survey}
Saurabh Arora and al.,
\newblock ``A survey of inverse reinforcement learning: Challenges, methods and
  progress,'' 2020.

\bibitem{adams2022survey}
Stephen Adams and al.,
\newblock ``A survey of inverse reinforcement learning,''
\newblock {\em Artif. Intell. Rev.}, vol. 55, no. 6, pp. 4307–4346, aug 2022.

\bibitem{Cleveland1988}
W.S. Cleveland and S.J. Devlin,
\newblock ``Locally weighted regression: An approach to regression analysis by
  local fitting,''
\newblock {\em Journal of the American Statistical Association}, vol. 83, no.
  403, pp. 596--610, 1988.

\bibitem{pari2021surprising}
Jyothish Pari and al.,
\newblock ``The surprising effectiveness of representation learning for visual
  imitation,'' 2021.

\bibitem{haldar2023teach}
Siddhant Haldar and al.,
\newblock ``Teach a robot to fish: Versatile imitation from one minute of
  demonstrations,'' 2023.

\bibitem{bahl2023affordances}
Shikhar Bahl and al.,
\newblock ``Affordances from human videos as a versatile representation for
  robotics,'' 2023.

\bibitem{ShahDragan2021}
Rohin Shah and al.,
\newblock ``The minerl {BASALT} competition on learning from human feedback,''
\newblock {\em CoRR}, vol. abs/2107.01969, 2021.

\bibitem{Milani2023}
Stephanie Milani and al.,
\newblock ``Towards solving fuzzy tasks with human feedback: A retrospective of
  the minerl basalt 2022 competition,'' 2023.

\bibitem{baker2022video}
Bowen Baker and al.,
\newblock ``Video pretraining (vpt): Learning to act by watching unlabeled
  online videos,''
\newblock {\em Advances in Neural Information Processing Systems}, vol. 35, pp.
  24639--24654, 2022.

\bibitem{EspeholtKavukcuoglu2018}
Lasse Espeholt and al.,
\newblock ``{IMPALA:} scalable distributed deep-rl with importance weighted
  actor-learner architectures,''
\newblock {\em CoRR}, vol. abs/1802.01561, 2018.

\bibitem{VPTGithub}
``Video-pre-training,''
  \url{https://github.com/openai/Video-Pre-Training/tree/main/lib},
\newblock Accessed: 2023-08-16.

\bibitem{minka2018trueskill}
Tom Minka and al.,
\newblock ``Trueskill 2: An improved bayesian skill rating system,''
\newblock Tech. {R}ep. MSR-TR-2018-8, Microsoft, March 2018.

\bibitem{faiss}
``Faiss: a library for efficient similarity search and clustering of dense
  vectors,'' https://ai.meta.com/tools/faiss/,
\newblock Accessed: 06-09-2023.

\end{thebibliography}

\end{document}